\RequirePackage{fix-cm}
\documentclass[twocolumn]{svjour3}          
\smartqed  
\usepackage{graphicx}
\usepackage{mathptmx}
\usepackage{amsfonts}
\usepackage{amsmath}
\usepackage{balance}
\usepackage{color}
\usepackage{booktabs}
\usepackage[table]{xcolor}

\newcommand{\TA}[1]{\cellcolor[rgb]{0.5,1.0,0.5}{\color{black} {\bf #1}}} 
\newcommand{\TB}[1]{{\bf #1}}

\newcommand{{\hx}}  {\widehat{x}}

\newcommand{\ru}    {\rule{0mm}{2.75mm}}

\newcommand{\be}    {\begin{equation}}
\newcommand{\ee}    {\end{equation}}

\begin{document}
\title{Combining PRNU and noiseprint for robust and efficient device source identification}

\author{ Davide Cozzolino \and Francesco Marra \and Diego Gragnaniello  \and Giovanni Poggi \and Luisa Verdoliva}

\institute{ \mailname{ verdoliv@unina.it} \\
              University Federico II of Naples \\
              Via Claudio, 21, 80125 Naples (Italy) \\
              Tel.: +39 081 76-83929\\
}


\maketitle

\begin{abstract}
PRNU-based image processing is a key asset in digital multimedia forensics.
It allows for reliable device identification and effective detection and localization of image forgeries, in very general conditions.
However, performance impairs significantly in challenging conditions involving low quality and quantity of data.
These include working on compressed and cropped images, or estimating the camera PRNU pattern based on only a few images.
To boost the performance of PRNU-based analyses in such conditions we propose to leverage the image noiseprint,
a recently proposed camera-model fingerprint that has proved effective for several forensic tasks.
Numerical experiments on datasets widely used for source identification
prove that the proposed method ensures a significant performance improvement in a wide range of challenging situations.
\end{abstract}

\section{Introduction}
\label{sec:introduction}

Image attribution is a fundamental task in the context of digital multimedia forensics
and has gained a great deal of attention in recent years.
Being able to identify the specific device that acquired a given image
represents a powerful tool in the hands of investigators fighting such hideous crimes as terrorism and pedopornography.

Most successful techniques for device identification rely on the photo response non-uniformity (PRNU) \cite{Lukas2006,Chen2008}, a sort of fingerprint left by the camera in each acquired photo.
The PRNU pattern originates from subtle random imperfections of the camera sensor which affect in a deterministic way all acquired images.
Each device has its specific PRNU pattern, which can be accurately estimated by means of sophisticated processing steps,
provided a large number of images acquired by the device itself is available.
Given the camera PRNU pattern, image attribution is relatively easy and very reliable, in ideal conditions.
However, the performance degrades rapidly when the operating conditions are less favourable, since it is ultimately related to the quantity of available data and their quality \cite{Rosenfeld2009}.
In particular, performance may be severely impaired when
{\it  i)} tests take place on small image crops,
{\it ii)} a limited number of images are available for estimating the device PRNU.
These challenging cases are of great interest for real-world applications, as clarified in the following.

Modern cameras generate very large images, with millions of pixels, which provide abundant information for reliable device identification.
However, often large-scale investigations are necessary.
The analyst may be required to process a large number of images, for example all images downloaded from a social account, and look for their provenance in a huge database of available cameras (PRNU patterns).
This calls for an inordinate processing time, unless suitable methods are used to reduce computation, typically involving some forms of data summarization
\cite{Bayram2012,Valsesia2015,Bondi2019}.
Working on small crops, rather than on the whole image, is a simple and effective way to achieve such a goal.
In addition,
the PRNU pattern can be also used to perform image forgery localization \cite{Chen2008,Chierchia2014}.
In this case, a PRNU-based sliding-window analysis is necessary, on patches of relatively small size.

Turning to the second problem,
it may easily happen, in criminal investigations, that only a few images acquired by the camera of interest are available to estimate its PRNU pattern.
Moreover, the number and type of source cameras may be themselves unknown, and must be estimated in a blind fashion based on a relatively small set of unlabeled images \cite{Bloy2008,Marra2017,Lin2016,Phan2018}.
In such cases, one must cope with low-quality estimation rather than abandoning the effort.

In this work, we tackle these problems and propose a novel source identification strategy
which improves the performance of PRNU-based methods when only a few, or even just one image is available for estimation, and when only small images may be processed.
To this end, we rely on a recent approach for camera model identification \cite{Cozzolino2019} and use it to improve the PRNU-based source device identification performance.
Camera model identification has received great attention in recent years, with a steady improvement of performance,
thanks to the availability of huge datasets on which it is possible to train learning-based detectors, and the introduction of convolutional neural networks (CNN).
The supervised setting guarantees very good performance \cite{Tuama2016}\cite{Obregon2018}, especially if deep networks are used.
However, such solutions are highly vulnerable to attacks \cite{Guera2017}\cite{Marra2018}.
To gain higher robustness, unsupervised or semi-supervised methods may be used.
For example, in \cite{Bondi2017} features are extracted through a CNN, while classification relies on machine learning methods.
Interestingly, only the classification step needs to be re-trained when testing on camera models that are not present in the training set.
Likewise, in \cite{Owen2018} it has been shown that proper fine-tuning strategies can be applied to camera model identification, a task that shares many features with other forensic tasks.
Of course, this makes the problem easier to face, given that in a realistic scenario it is not possible to include in the training phase all the possible camera models.
A further step in this direction can be found in \cite{Cozzolino2019},
where the use of a new fingerprint has been proposed, called noiseprint, related to camera model artifacts and extracted by means of a CNN trained in siamese modality.
Noiseprints can be used in PRNU-like scenarios but require much less data to reach a satisfactory performance \cite{Cozzolino2018}.

The main idea of this paper is to make use of noiseprints to support PRNU-based device identification.
In fact, although noiseprints allow only for model identification, they are much stronger than PRNU patterns, and more resilient to challenging conditions
involving restrictions on the number and size of images.

In the rest of the paper,
after providing background material on PRNU and noiseprint in Section II,
and describing the proposed method in Section III,
we carry out a thorough performance analysis, in Section IV, on datasets widely used in the literature and in several conditions of interest, proving the potential of the proposed approach.
Finally, Section V concludes the paper.

\section{Background}
\label{sec:background}

Device-level source identification relies on specific marks that each individual device leaves on the acquired images.
Barring trivial cases, like the presence of multiple defective pixels in the sensor,
most identification methods resort to the photo response non-uniformity noise (PRNU) pattern.
In fact, due to unavoidable inaccuracies in sensor manufacturing, sensor cells are not perfectly uniform and generate pixels with slightly different luminance in the presence of the same light intensity.
Accordingly, a simplified multiplicative model for an image $I$ generated by a given camera is
\begin{equation}
    I = (1+K)I^0 + \Theta
\end{equation}
where $I^0$ is the true image, $K$ is the PRNU pattern, $\Theta$ accounts for all other sources of noise, and all operations are pixel-wise.
The PRNU is unique for each device, stable in time, and present in all images acquired by the device itself.
Therefore it can be regarded as a legitimate device fingerprint, and used to perform a large number of forensic tasks \cite{Chen2008,Goljan2008,Chierchia2014}.

The PRNU of camera $C_i$ can be estimated from a suitable number of images, say, $I_{i,1},\ldots,I_{i,N}$, acquired by the camera itself.
First, noise residuals are computed by means of some suitable denoising algorithms $f(\cdot)$
\begin{equation}
    W_{i,n} = I_{i,n}-f(I_{i,n})
\end{equation}
The denoiser removes to a large extent the original image, $I^0_{i,n}$, regarded as a disturbance here, in order to emphasize the multiplicative pattern \cite{Chierchia2010}.
Then the PRNU, $K_i$, can be estimated by plain average
\begin{equation}
    \widehat{K}_i = \frac{1}{N} \sum_{n=1}^N W_{i,n}
\end{equation}
or according to a maximum-likelihood rule, in order to reduce the residual noise.
Moreover, to remove unwanted periodicities, related to JPEG compression or model-based artifacts,
the estimated fingerprint is further filtered by subtracting the averages of each row and column and by using a Wiener filter in the DFT domain \cite{Lukas2006,Wiener1964}.
Eventually, the estimate converges to the true PRNU as $N \to \infty$.

Assuming to know the true reference PRNU, we can check if image $I_m$ was acquired by camera $C_i$ based on the normalized cross-correlation (NCC) between $W_m$ and $K_i$
\begin{equation}
    {NCC}(W_m,K_i) = \frac{1}{\|W_m\|\cdot\|K_i\|} \langle W_m,K_i \rangle
\end{equation}
with $\langle \cdot,\cdot \rangle$ and $\|\cdot\|$ indicating inner product and Euclidean norm, respectively.
The computed NCC will be a random variable, due to all kinds of disturbances, with a positive mean if the image was taken by camera $C_i$, and zero mean otherwise.
In the following, to allow easier interpretation, we present results in terms of a PRNU-based pseudo- distance, defined as the complement to 1 of NCC
\begin{equation}
    D_{\rm PRNU}(i,m) = 1 - NCC(W_m,K_i)
\end{equation}

Methods based on PRNU have shown excellent per\-for\-mance for source identification \cite{Chen2008} in ideal conditions.
However, the NCC variance increases both with decreasing image size and when the number of images used to estimate the real PRNU decreases, in which cases, decision may become unreliable.
Both conditions may easily occur in real-world practice, as recalled in the Introduction.
Therefore, further evidence, coming from camera-model features, is highly welcome to improve identification performance.

Indeed, it is well known that, besides device-specific tra\-ces, each image bears also model-specific traces, related to the processes carried inside the camera, such as demosaicking or JPEG compression.
Such traces are treated as noise in the PRNU estimation process, and mostly removed, but they can provide a precious help to tell apart cameras of different models.

\begin{figure}
	\centering
	\includegraphics[width=1.00\linewidth]{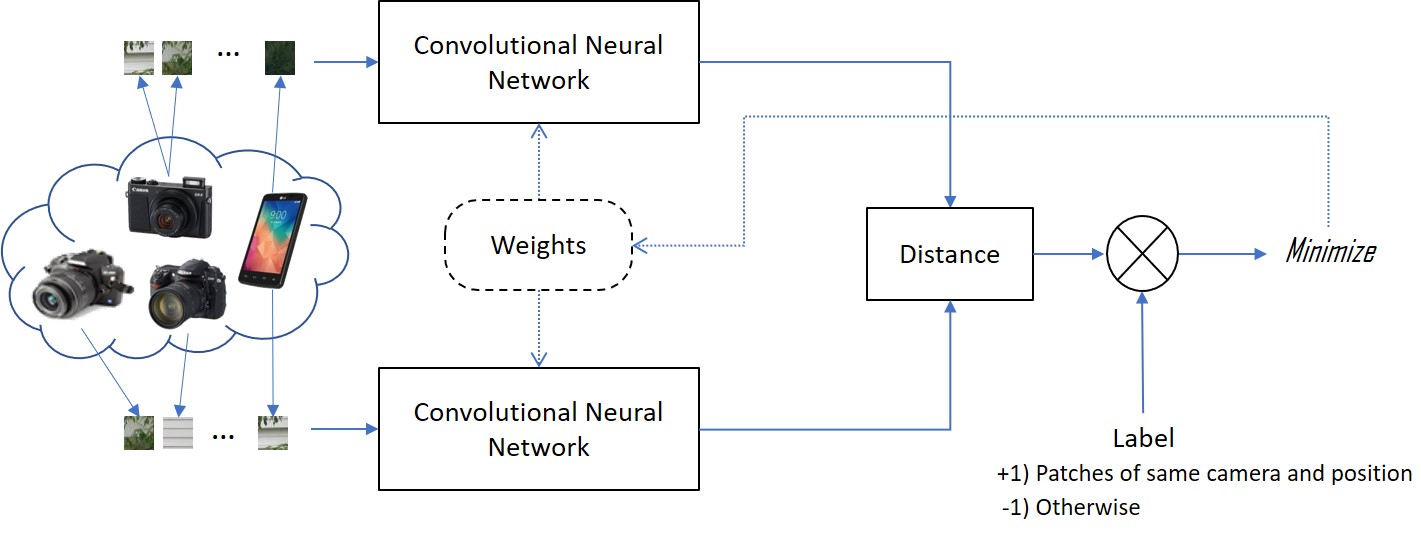}
	\caption{Using a Siamese architecture for training.
		The output of one CNN takes the role of desired (same model and position) or undesired (different models or positions) reference for the other twin CNN.}
	\label{fig:noiseprint_training}
\end{figure}

Recently, a camera-model fingerprint has been proposed \cite{Cozzolino2019}, called noiseprint, extracted by means of a suitably trained convolutional neural network (CNN).
A noiseprint is an im\-age-size pattern, like the PRNU, in which the high-level scene content is removed and model-related artifacts are emphasized.
This is obtained by training the CNN in a Siamese configuration, as illustrated in Fig.\ref{fig:noiseprint_training}.
Two identical versions of the same net are fed with pairs of patches extracted from unrelated images.
Such pairs have positive label when they come from the same model and have the same spatial position, and negative label otherwise.
During training, thanks to a suitable loss function,
the network learns to emphasize the similarities among positive pairs, that is, the camera model artifacts.
In \cite{Cozzolino2019} noiseprints have been shown to enable the accomplishment of numerous forensic tasks, especially image forgery localization.

Similarly to the reference PRNU pattern,
the reference noiseprint of a model is obtained by averaging a large number of noiseprints extracted by images acquired by the same model (not necessarily the same device) \cite{Cozzolino2018},
in formulas
\begin{equation}
    R_i = \frac{1}{N} \sum_{n=1}^N \phi(I_{i,n})
\end{equation}
where $\phi(\cdot)$ is the function implemented by the CNN, $R_i$ is the estimated reference pattern of the $i$-th model and, $I_{i,n}$ is the $n$-th image taken by the $i$-th model.
To verify whether a given image $I_m$ was acquired by camera model $M_i$, we extract its noiseprint, $\phi(I_m)$, and compute the mean square error (MSE) with respect to the model reference pattern $R_i$.
For homogeneity with the previous case, this is also called {\em NP-based} distance
\begin{equation}
    D_{\rm NP}(i, m) = {\rm MSE}(R_i, \phi(I_m))
\end{equation}
Again $D_{\rm NP}(i, m)$ is expected to be small if the $m$-th image was indeed acquired by the $i$-th model and large otherwise,
and its variance depends, again, on the size and number of images used to estimate the reference pattern.

\begin{figure}
    \centering
    \includegraphics[width=0.9\linewidth]{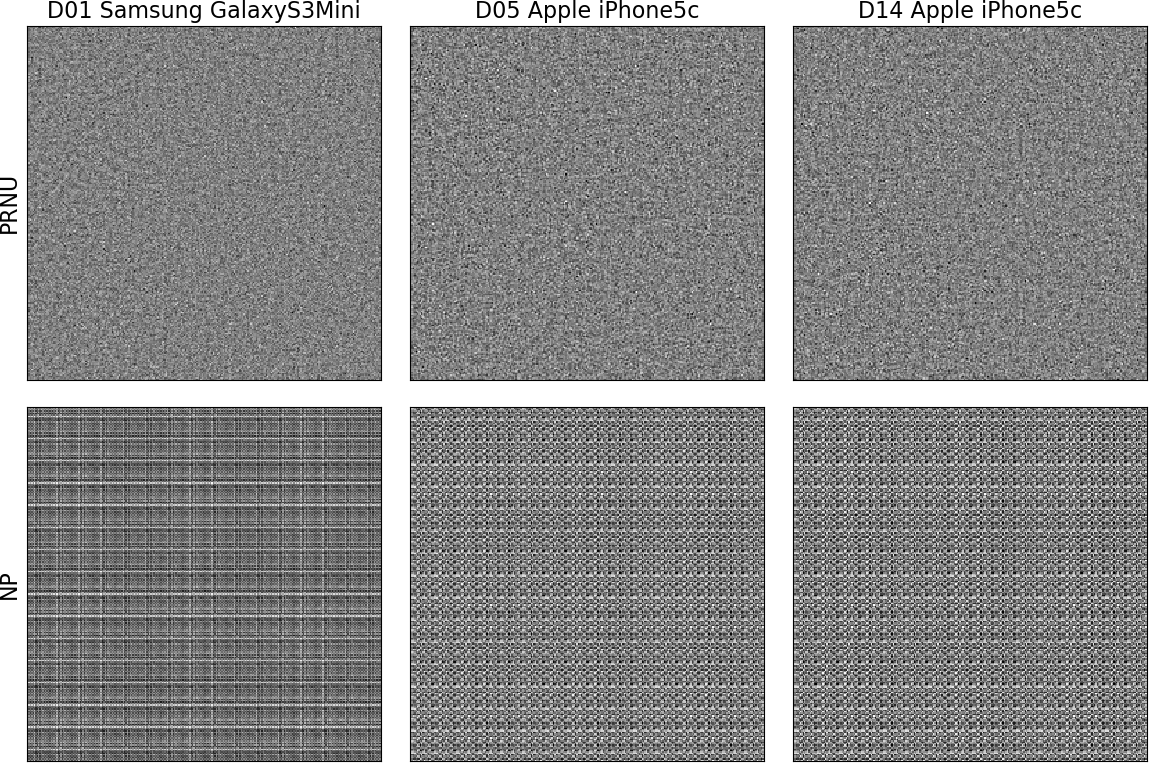}
    \caption{Examples of device fingerprints (PRNU, top) and model fingerprints (noiseprint, bottom) for cameras of the VISION dataset,
    estimated using 100 images on a crop of dimension 256$\times$256.
    Device fingerprints are noise-like, weak patterns, all different from one another.
    Noiseprints are much stronger, model-specific, periodic patterns.
    Note that the last two devices are of the same model and, accordingly, their noiseprints are almost identical.}
    \label{fig:PRNU-vs-NP}
\end{figure}

While the image noiseprint does not allow to single out the camera that acquired a given image,
it can be used to discard or play down a large number of candidates, that is, all cameras whose model does not fit the test noiseprint.
Moreover, the model-related traces found in noiseprints are much stronger than the PRNU traces found in image residuals,
as clearly shown in the example of Fig.\ref{fig:PRNU-vs-NP},
hence noiseprint-based decisions keep being reliable even over small regions and when a limited number of images is available for estimation \cite{Cozzolino2018}.

\section{Proposed method}
\label{sec:method}

In this work, we want to improve the performance of the device source identification in the two critical scenarios:
{\it  i)} only a few images are available to estimate the reference PRNU pattern; and
{\it ii)} only a small crop of the image is used for testing.
The few-image scenario accounts for cases where the physical device is not available,
so that the only available images are recovered from a hard disk or maybe from a social account of a suspect.
Instead, the small-region scenario is motivated by the need to reduce memory and time resources for very large-scale analyses.
In addition, a good performance on small regions allows one to use this approach for image forgery detection and localization,
especially the critical case of small-size forgeries.

\begin{figure}
    \centering
    \includegraphics[width=\linewidth]{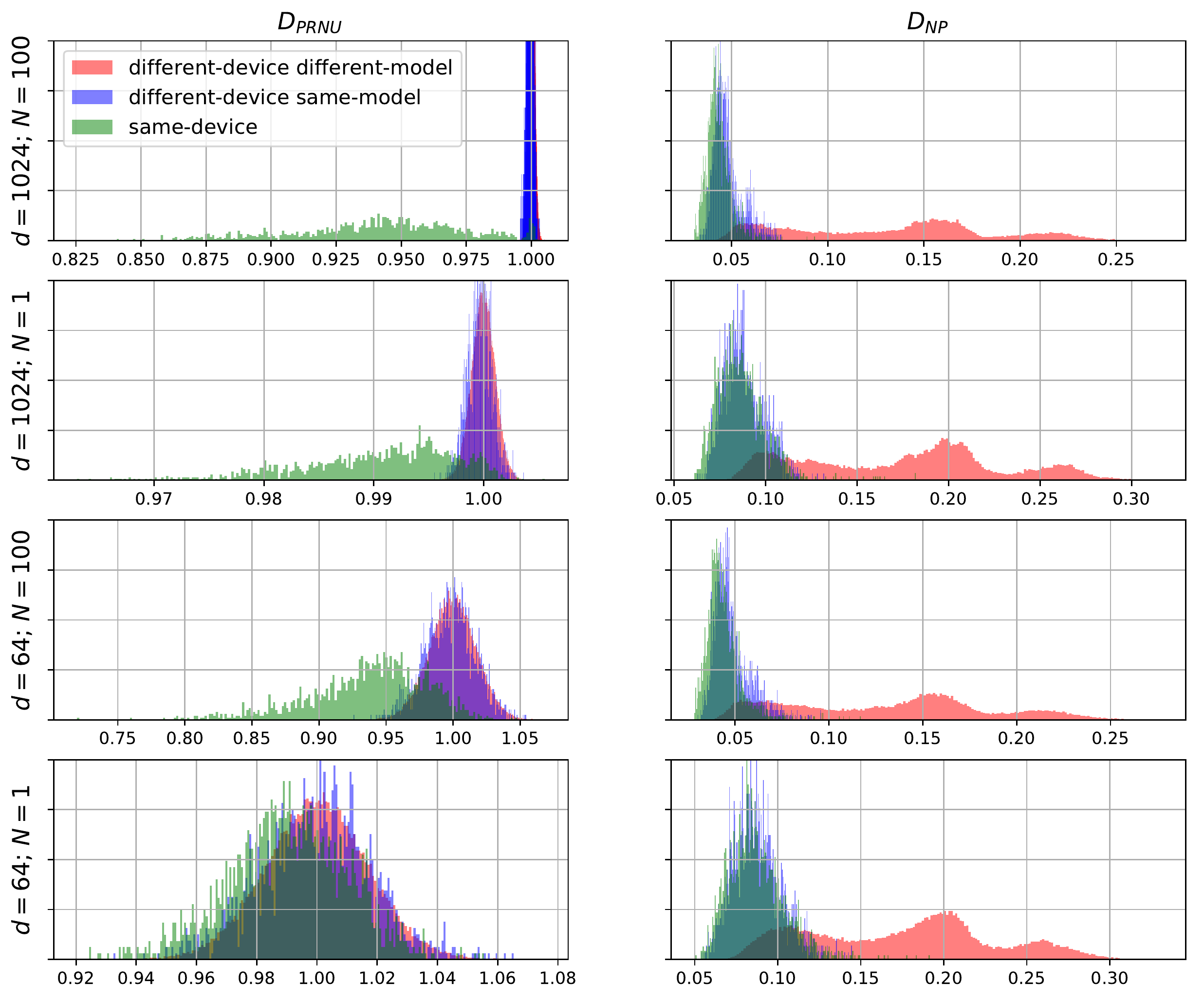}
    \caption{Histograms of PRNU-based distance, $D_{PRNU}$ (left), and noiseprint-based distance, $D_{NP}$ (right), on the VISION dataset.
    $N$: number of images used to estimate the reference pattern. $d \times d$: size of crop used for testing.}
    \label{fig:histograms_vision}
\end{figure}

In Fig.\ref{fig:histograms_vision}, we show the histograms of the PRNU-based (left) and noiseprint-based (right) distances
computed on the widespread VISION dataset \cite{Shullani2017} in various situations (different rows).
Each subplot shows three different histograms:
\begin{enumerate}
    \item same-device (green):                    the distance is evaluated between an image acquired from camera $C$ of model $M$ and the reference pattern of the same camera;
    \item different-device same-model (blue):     the distance is evaluated between an image acquired from camera $C$ of model $M$ and the reference pattern of a different camera $C'$ of the same model $M$;
    \item different-device different-model (red): the distance is evaluated between an image acquired from camera $C$ of model $M$ and the reference pattern of a different camera $C'$ of a different model $M'$.
\end{enumerate}
On the first row, we consider a nearly ideal situation,
where tests are performed on 1024$\times$1024 crops of the image and there is plenty of images (we limit them to $N$=100) to estimate the reference patterns.
The subplot on the left shows that the PRNU-based distance separates very well same-de\-vice from different-device samples.
On the contrary, the two different-device distributions (same-model and different- model) overlap largely,  since the PRNU does bear model-specific information.
Then, when the number of images used to estimate the reference pattern decreases (second row, $N$=1)
or the analysis crop shrinks significantly (third row, $d$=64)
same-device and different-device distributions are not so well separated anymore, and PRNU-based decisions become unreliable.
Eventually, in the extreme case of $N$=1 and $d$=64 (fourth row), all distributions collapse.
The right side of the figure shows histograms of the noiseprint-based distance.
In the ideal case (top row) we now observe a very good separation between same-model and different-model histograms,
while the two same-model distributions overlap, as noiseprints do not carry device-related information.
Unlike with PRNU, however,
when the analysis conditions deviate from ideal (following rows), the same-model and different-model distributions keep being reasonably well-separated,
allowing for a reliable model discrimination even in the worst case (fourth row).
This suggests the opportunity to use the noiseprint-based distance to support decision insofar different models are involved,

\begin{figure*}
    \centering
    \includegraphics[width=0.9\linewidth]{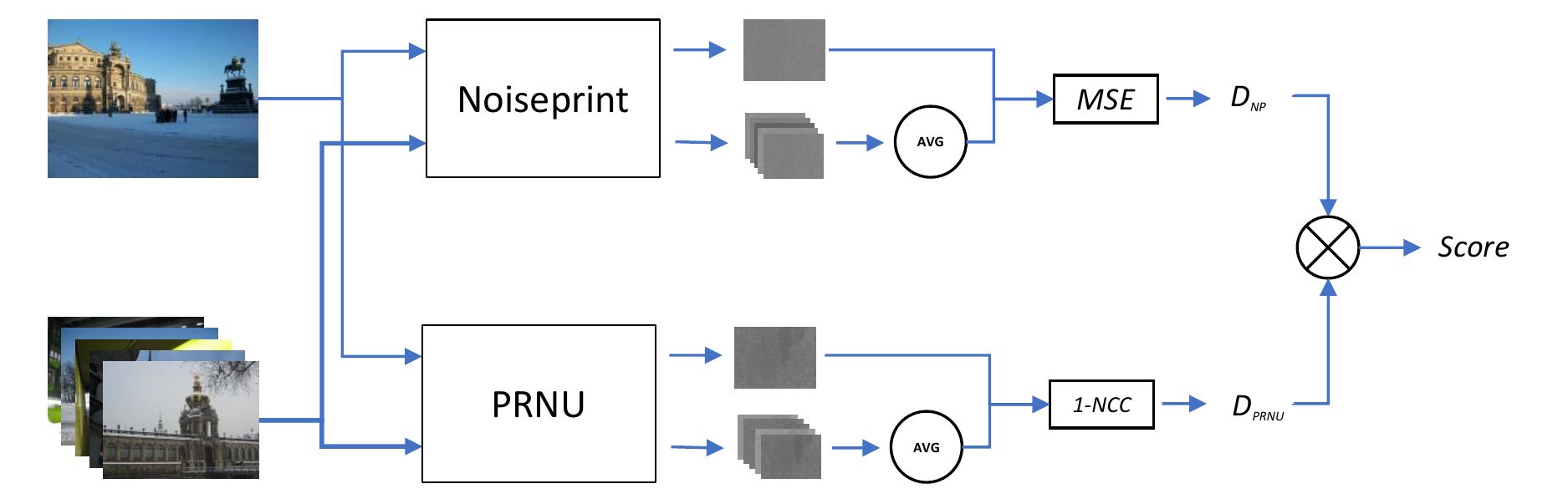}
    \caption{The PRNU and the noiseprint are estimated from available images of a given camera and the image under test, separately. Thus, the PRNU-based ($D_{PRNU}$) and the noiseprint-based ($D_{NP}$) pseudo-distances between the test sample and the reference are separately computed and eventually combined to obtain the final score.}
    \label{fig:schema}
\end{figure*}

In summary,
our proposal is to use noiseprint-based mo- del-related information to support PRNU-based device identification.
Assuming to know the camera model of the image to analyze,
the search for the source device can be restricted only to devices of the same model, thereby reducing the risk of wrong identification, especially in the most critical cases.
However, in real-world scenarios, camera models may not be known in advance,
calling for a preliminary model identification phase, which is itself prone to errors.
Therefore, with the hierarchical procedure outlined before, there is the non-negligible risk of excluding right away the correct device.
For this reason, we prefer to exploit the two pieces of information jointly, rather than hierarchically, by suitably combining the two distances,
as shown pictorially in Fig.\ref{fig:schema}.
Note that, in this worst-case scenario, where no model-related information is known a priori,
the noiseprint reference pattern is estimated from the very same images used to estimate the PRNU pattern.
Hence, the resulting performance represents a lower bound for more favourable conditions.

We consider the following binary hypothesis test:
\begin{equation}
    \begin{cases}
    H_0: & \mbox{$m$-th image {\em not} acquired by $i$-th device.} \\
    H_1: & \mbox{$m$-th image           acquired by $i$-th device.}
    \end{cases}
    \label{eq:hypothesis}
\end{equation}
and propose three different strategies to combine the two distances.

\paragraph{SVM:}
as first strategy, we adopt a linear support-vector machine (SVM) classifier.
Relying on a large dataset of examples, the SVM finds the hyperplane that best separates samples of the two hypotheses, maximizing the distance from the hyperplane to the nearest points of each class.
Then, we use the oriented distance from the hyperplane as a score to detect the correct device that acquired the image.
If a limited number of candidate devices is given, the maximum-score device is chosen.
Otherwise, the score is compared with a threshold to make a binary decision, and the threshold itself may be varied to obtain a performance curve.
These criteria apply to the following cases as well.

\paragraph{Likelihood-Ratio Test:}
to formulate a likelihood-ratio test (LRT), we model $D_{\rm PRNU}$ and $D_{\rm NP}$ as random variables with jointly Gaussian distribution in both hypotheses
\begin{equation}
    \begin{cases}
    H_0: & \left( D_{PRNU}, D_{NP} \right) \sim {\cal N}\left( \mu_0, \Sigma_0 \right) \\
    H_1: & \left( D_{PRNU}, D_{NP} \right) \sim {\cal N}\left( \mu_1, \Sigma_1 \right)
    \end{cases}
    \label{eq:hypothesis}
\end{equation}
The parameters $\mu_0$, $\Sigma_0$, $\mu_1$, $\Sigma_1$ are estimated, for each situation of interest, on a separate training-set.
In particular, we use two methods to estimate these parameters,
the classical ML approach, and a robust approach based on the minimum covariance determinant MCD proposed in \cite{Rousseeuw1999}.
With this Gaussian model, and the estimated parameters, we can compute the log-likelihood ratio
\begin{equation}
    \Lambda(i, m) = \log\frac{  {\cal N}\left( D_{PRNU}(i,m), D_{NP}(i,m) \left | \mu_1, \Sigma_1 \right. \right) }
                             {  {\cal N}\left( D_{PRNU}(i,m), D_{NP}(i,m) \left | \mu_0, \Sigma_0 \right. \right) }
    \label{eq:lrt}
\end{equation}
and use it as decision statistic.

\paragraph{Fisher's Linear Discriminant:}
this approach looks for the direction along which the distributions of two classes are better separated, measuring the separation as the ratio of the inter-class to intra-class variances.
The weight vector of the optimal direction, $w^{\rm opt}$, is given by
\begin{equation}
    w^{\rm opt}  = \arg\max_w \frac{\left( w^T \mu_1 - w^T \mu_0 \right)^2}{w^T \left(\Sigma_1 + \Sigma_0 \right) w}
                   \propto \left(\Sigma_1 + \Sigma_0 \right)^{-1} \,\, \hspace{-2mm} \left(\mu_1 - \mu_0 \right)
\end{equation}
where, again, means and co-variance matrices in the two hypotheses, $\mu_0$, $\Sigma_0$, $\mu_1$, $\Sigma_1$, are estimated on a training-set.
In this case, the score is given by the projection of the vector formed by the two distances along the optimal direction
\begin{equation}
    {FLD}(i, m) = w^{\rm opt}_{PRNU}D_{PRNU}(i, m) +  w^{\rm opt}_{NP}D_{NP}(i, m)
    \label{eq:linear}
\end{equation}

\begin{table}[!ht]
	\centering
	\begin{footnotesize}
		\begin{tabular}{llrrr}
			\toprule
			Make	& Model		    & \#Devices	&	Images size	\\ \toprule
			Canon	& Ixus 70	    &	3		&	3072 x 2304	\\
			Casio	& EX-Z150	    &	5		&	3264 x 2448	\\
			Kodak	& M1063		    &	5		&	3664 x 2748	\\
			FujiFilm& FinePixJ50    &	3		&   3264 x 2448 \\
			Nikon	& CoolPix S710	&	5	    &	4352 x 3264	\\
	        Nikon 	& D200		    &	2		&	3872 x 2592	\\
	        Nikon	& D70/D70s		&	2/2		&	3008 x 2000	\\
			Olympus	& $\mu$1050SW	&	5	    &	3648 x 2736	\\
			Panasonic	& DMC-FZ50	&	3		&	3648 x 2736	\\
			Pentax	& Optio A40	    &	4		&	4000 x 3000	\\
			Pratika	& DCZ5.9	    &	5		&	2560 x 1920	\\
			Ricoh	& GX100	        &	5	    &	3648 x 2736	\\
			Rollei	& RCP-7325XS	&	3		&	3072 x 2304	\\
			Samsung	& L74wide	    &	3		&	3072 x 2304	\\
			Samsung	& NV15		    &	3		&	3648 x 2736	\\
			Sony	& DSC-H50		&	2		&	3456 x 2592 \\
			Sony	& DSC-T77		&	4		&	3648 x 2736 \\
			Sony	& DSC-W170		&	2		&	3648 x 2736 \\
			\midrule
			$\Sigma$	& 18 	    &  66 		& 				\\
			\bottomrule
		\end{tabular}
	\end{footnotesize}
	\caption{Camera models and devices selected from the Dresden Image Database used for testing.  For each camera, we considered 100 images to estimate the reference patterns and 50 images   for testing.}
	\label{tab:DresdenCameras}
\end{table}

\section{Experiments}
In this section, we assess the performance of the proposed device identification method in all its variants,
and in various realistic scenarios of interests.
We consider two typical scenarios (Fig.\ref{fig:closed-open}):

\begin{figure}
    \centering
    \includegraphics[width=1.0\linewidth]{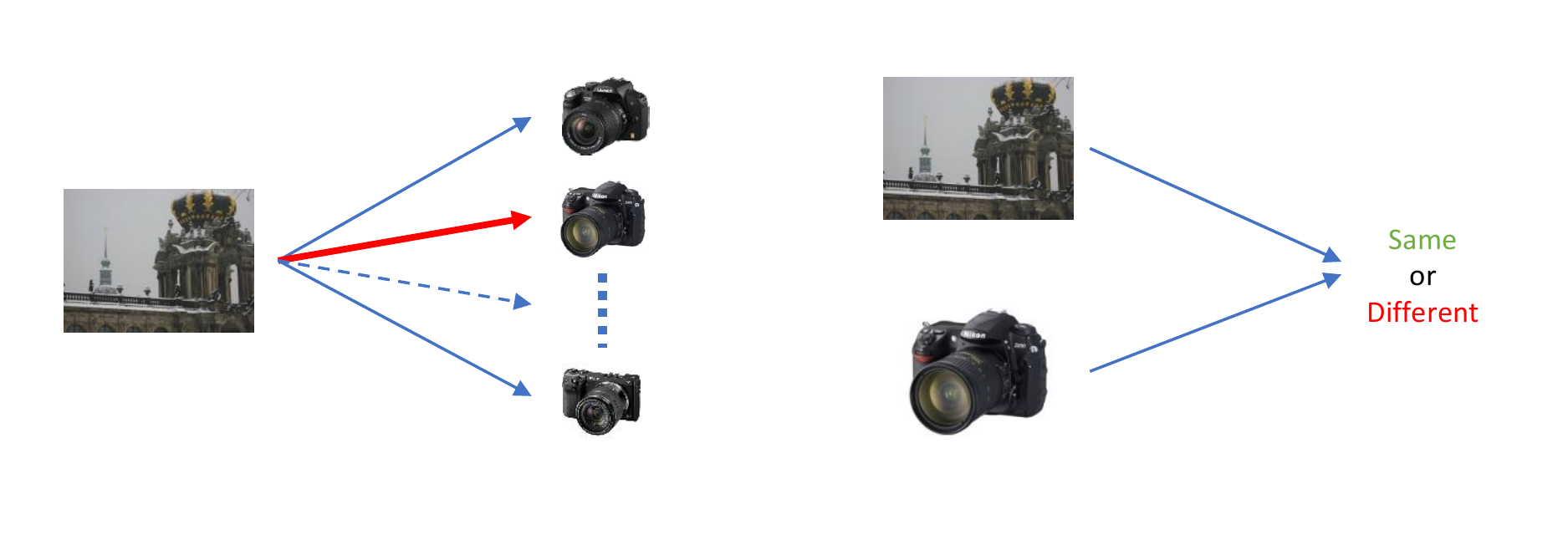}
    \caption{On the left the \emph{closed-set} scenario where the test image is assigned to one of the device in a set of known cameras.
    In the \emph{verification} or \emph{open-set} scenario, on the right,
    the task is to decide if the test image was acquired by a certain device or not.}
    \label{fig:closed-open}
\end{figure}

\begin{enumerate}
    \item \emph{closed-set}.
    The set of candidate sources is predefined and one is required to associate the test image with one of the devices in the set.
	Therefore, the task can be regarded as a multi-class classification problem.
    Accordingly, we evaluate the classification performance in terms of accuracy, that is, probability of correct decision.
	
	\item \emph{open-set} or \emph{verification}.
    The set of candidate sources is not given a priori.
	In this case, one can only decide on whether the test image was acquired by a certain device or not.
	Therefore, the problem is now binary classification,
    and we evaluate the classification performance, as customary, in terms of probability of correct detection $P_D$ and probability of false alarm $P_{FA}$,
    summarized by a receiver operating curve (ROC), computed for varying decision threshold,
    and eventually by the area under such a curve (AUC).
\end{enumerate}

\subsection{Datasets}
Three different datasets have been employed for our experiments, two for training and one for testing.
In fact, in order to avoid any bias, we trained the CNN that extracts the noiseprints and the source identification classifiers on disjoint sets of data.
Concerning noiseprints,
the network is trained on a large dataset of images publicly available on dpreviewer.com.
Our collection comprises 625 camera models, 600 for training the network and 25 for validation,
with a number of images per model ranging from 8 to 173.
The parameters of the source identification classifiers, instead, have been estimated on the VISION dataset \cite{Shullani2017},
often used for camera model identification, which comprises images acquired from 29 different models, most of them with only one device.
Finally, tests have been conducted on the Dresden dataset \cite{Gloe2010}, proposed originally for camera source identification,
comprising images from 25 camera models, 18 of which featuring two or more different devices.
We selected these latter models with all their devices, for a grand total of 66.
Details are reported in Tab.\ref{tab:DresdenCameras}.

\begin{figure}
    \centering
    \includegraphics[width=\linewidth]{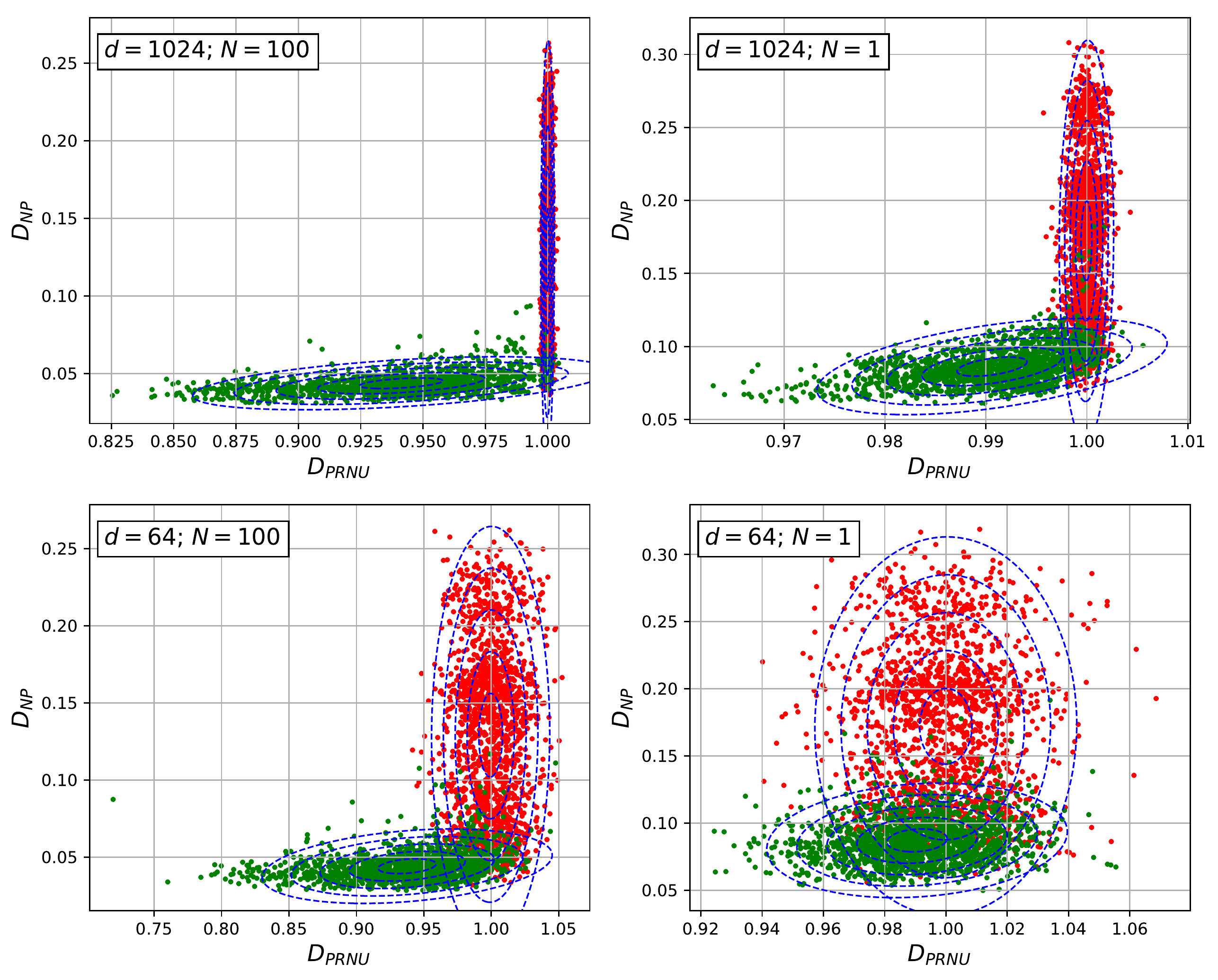}
    \caption{Scatter plots of the two distances in different conditions on the VISION dataset.
    For green points, test image and reference come from the same device, while for red points, they come from different devices.
    Blue lines depict the Gaussian fitting of the two distributions.}
    \label{fig:scatters_vision}
\end{figure}

\subsection{Results}
To assess the source identification performance of the conventional PRNU-only and the proposed PRNU+noiseprint methods we consider several challenging cases
obtained by varying the size of the image crop used for testing and the number of images used for estimating the reference patterns.
In addition, we consider also the case of JPEG images compressed at two quality factors,
aimed at simulating a scenario in which images are downloaded from social network accounts, where compression and resizing are routinely performed.
Note that the PRNU and noiseprint reference patterns are both estimated from the very same images, since no prior information is assumed to be available on the camera models.

To gain some insight into the role of the two distances,
Fig.\ref{fig:scatters_vision} shows the scatter plots of $D_{PRNU}$ and $D_{PN}$ for images of the VISION dataset in the same cases of Fig.\ref{fig:histograms_vision}.
Here, however, we only show same-device (green) and different-device (red) points, irrespective of models.
In the most favourable case of $d$=1024 and $N$=100 (top-left),
the two clusters can be separated very effectively by a vertical line, that is, based only on the PRNU distance.
Instead, in the worst case of $d$=64 and $N$=1 (bottom-right),
the PRNU-based distance is basically useless, while the noiseprint-based distance provides by itself a pretty good separation,
with residual errors mostly due to same-model different-device points.
In intermediate cases, both pieces of information help separating effectively the two clusters.
Note that a weak correlation between the two distances exist for same-device points,
likely due to the imperfect rejection of unwanted contributions.
Of course, a different classifier must be designed for each situation of interest, based on the corresponding training set.

We now analyze the performance on the test dataset, Dresden, never used in training,
beginning with the closed-set scenario, Tab.\ref{tab:ACC}.
On the rows, we consider all combinations of $d$=64, 256, 1024, and $N$=1, 10, 100.
When $N$=1 or 10, results are averaged over 10 repetitions with different reference patterns.
The leftmost columns show results for the two baselines, based only on the PRNU-distance (that is, the conventional method of \cite{Chen2008}) and only on the noiseprint-distance.
Again, the performance of the conventional method is very good in the ideal case,
but worsens significantly in more challenging situations, and is only slightly better than random choice when $N$=1 and $d$=64.
On the contrary, noiseprint-only identification is never very good, since there are always from 2 to 5 indistinguishable devices for each model,
but remains remarkably stable in all conditions,
suggesting that models keep being identified accurately, with only minor impairments even in the most challenging cases.
In the following column we show a third reference, called ``ideal'', which represents an upper bound for the performance of the proposed approach.
Here, we assume perfect model identification,
and rely on the PRNU-based distance only for the final choice in the restricted set of same-model cameras.
In the most favourable case ($d$=1024, $N$=100) the PRNU-only performance was already quite good, and only a marginal improvement is observed.
Likewise, in the worst case ($d$=64, $N$=1), PRNU is unreliable, and the fusion improves only marginally upon the noiseprint-only performance.
However, in intermediate cases, large improvements are observed, with the accuracy growing from 0.649 to 0.793 (case $d$=1024, $N$=1) or from 0.342 to 0.610 ($d$=64, $N$=100).
This performance gain fully justifies our interest for this approach,
we only need the proposed methods to work relatively close to this upper bound.

\begin{table*}[!t]
	\setlength\tabcolsep{6pt}
	\renewcommand{\ru}{\rule{0mm}{3mm}}
	\centering
	\footnotesize
	\begin{tabular}{rr||c|c||c||c|c|c|c|c}
		\ru    d &   N & PRNU  &  NP   & ideal &      SVM   &      LRT   &     r-LRT  &      FLD   &     r-FLD  \\ \hline\hline
		\ru      &   1 & 0.044 & 0.233 & 0.331 &     0.255  &     0.240  &     0.238  & \TA{0.258} &     0.256  \\
		\ru   64 &  10 & 0.117 & 0.259 & 0.422 & \TA{0.368} &     0.348  &     0.358  &     0.351  &     0.341  \\
		\ru      & 100 & 0.342 & 0.282 & 0.610 & \TA{0.570} &     0.557  &     0.565  &     0.563  &     0.557  \\ \hline
		\ru      &   1 & 0.191 & 0.261 & 0.485 &     0.405  &     0.397  &     0.408  & \TA{0.430} &     0.429  \\
		\ru  256 &  10 & 0.500 & 0.272 & 0.721 &     0.683  &     0.664  &     0.662  & \TA{0.692} &     0.689  \\
		\ru      & 100 & 0.814 & 0.289 & 0.904 &     0.881  &     0.868  &     0.867  & \TB{0.882} & \TB{0.882} \\ \hline
		\ru      &   1 & 0.649 & 0.270 & 0.793 &     0.532  &     0.733  &     0.735  &     0.751  & \TB{0.752} \\
		\ru 1024 &  10 & 0.901 & 0.294 & 0.952 &     0.888  &     0.910  &     0.907  &     0.913  & \TB{0.919} \\
		\ru      & 100 & 0.953 & 0.321 & 0.975 & \TB{0.955} &     0.954  &     0.954  &     0.952  &     0.953  \\ \hline
	\end{tabular}
	\caption{Accuracy on the Dresden dataset in the closed-set scenario without compression (random-choice accuracy is 0.016).
		Here and in next tables, the best fusion results are in boldface. When they improve by 20\% or more over PRNU, they are emphasized with a green background.
		Large improvements are observed when there is data scarcity (small $d$, small $N$).
		All classifiers have quite similar performances.}
	\label{tab:ACC}
\end{table*}

In the following five columns we report results for the various versions of the proposed method,
based on support vector machine classifier (SVM), likelihood ratio test with ML estimation (LRT) and robust estimation (r-LRT) of parameters,
and Fisher's linear discriminant in the same two versions (FLD and r-FLD).
Results are fully satisfactory.
Taking for example the FLD column, the gap with respect to the ideal reference remains quite small in all cases of interest and,
consequently, a large improvement is observed with respect to the conventional method.
This confirms that the noiseprint-based model classification remains largely successful in most conditions, and helps improving the overall performance.
As for the various versions of the proposed approach,
there seems to be no consistent winner, with FLD providing slightly better results, on the average, and SVM showing an isolated bad point ($d$=1024, $N$=1) maybe due to some overfitting.
In particular, the nonlinear decision boundary of the LRT criterion does not seem to ensure improvements over the linear boundaries of SVM and FLD,
confirming that the two classes are linearly well separated in the feature space.
The versions based on robust estimation of parameters performs on par or slightly worse than the ML counterparts.

Tab.\ref{tab:AUC}, structurally identical to Tab.\ref{tab:ACC},
provides results for the open-set scenario in terms of area under the ROC curve (AUC).
All considerations made for the closed-set scenario keep holding here.
Of course, numbers are much larger than before, because we are considering binary decisions,
where an AUC of 0.5 is equivalent to coin tossing and good results correspond to AUC's close to 1.
This is actually the case for the proposed method.
Considering again the FLD column, the AUC is never less than 0.935 and not far from that of the ideal reference,
while it is often much larger than the AUC of the conventional method.
Actually, it is worth emphasizing that also the noiseprint-only method has quite good performance indicators.
In hindsight, this is not too surprising.
This scenario, in fact, fits the case of large-scale analysis, where a very large number of candidate sources must be considered.
Perfect {\em model} identification allows one to reject right away most of the candidates (small $P_{FA}$, large AUC),
which allows one to focus on a limited set of candidates, to be analyzed with greater care and resources.
Like before, differences among the various versions of the proposal are negligible.

\begin{table*}[]
	\centering
	\footnotesize
    \renewcommand{\ru}{\rule{0mm}{3mm}}
	\begin{tabular}{rr||c|c||c||c|c|c|c|c}
		\ru    d &   N & PRNU  &  NP   & ideal &      SVM   &      LRT   &     r-LRT  &      FLD   &     r-FLD  \\ \hline\hline
		\ru      &   1 & 0.580 & 0.937 & 0.980 &     0.938  &     0.932  &     0.932  & \TA{0.935} &     0.933  \\
		\ru   64 &  10 & 0.678 & 0.951 & 0.983 & \TA{0.952} &     0.951  & \TA{0.952} &     0.942  &     0.938  \\
		\ru      & 100 & 0.826 & 0.954 & 0.989 &     0.961  &     0.962  &     0.962  &     0.959  &     0.957  \\ \hline
		\ru      &   1 & 0.742 & 0.946 & 0.985 & \TA{0.949} &     0.945  &     0.946  &     0.945  &     0.944  \\
		\ru  256 &  10 & 0.880 & 0.956 & 0.992 &     0.965  &     0.966  &     0.966  & \TB{0.968} &     0.967  \\
		\ru      & 100 & 0.960 & 0.959 & 0.997 & \TB{0.980} &     0.978  &     0.978  &     0.979  &     0.979  \\ \hline
		\ru      &   1 & 0.906 & 0.949 & 0.994 &     0.952  &     0.964  &     0.964  &     0.964  & \TB{0.965} \\
		\ru 1024 &  10 & 0.981 & 0.958 & 0.999 &     0.971  &     0.981  &     0.979  &     0.980  & \TB{0.983} \\
		\ru      & 100 & 0.989 & 0.960 & 0.999 & \TB{0.984} &     0.983  &     0.983  &     0.982  &     0.983  \\ \hline
	\end{tabular}
    \caption{AUC on the Dresden dataset in the open-set scenario without compression (random-choice AUC is 0.5).
    Again, large improvements are observed in the presence of small patches and a small number of reference images.
    Also in the case, classifiers perform about equally well.}
	\label{tab:AUC}
\end{table*}

The next four tables refer to the case of images compressed using JPEG,
with QF=90 (Tab.\ref{tab:ACC_90} and Tab.\ref{tab:AUC_90}) and with QF=80 (Tab. \ref{tab:ACC_80} and Tab.\ref{tab:AUC_80})
always for both the closed-set and open-set scenarios.
First of all, with reference to the closed-set scenario, let us analyze the performance of the conventional method as the image quality impairs.
Only in the ideal case the accuracy is fully satisfactory,
while it decreases dramatically in all other conditions, for example, from 0.649 (uncompressed) to 0.364 (QF=80), for ($d$=1024, $N$=1).
In fact, the JPEG compression filters out as noise most of the small traces on which source identification methods rely.
This is also true for the noiseprint traces.
However, in the same case as before, with the robust-FLD version, the proposed method keeps granting an accuracy of 0.540,
with a more limited loss from the 0.752 accuracy of uncompressed images, and a large gain with respect to the conventional method.
The same behavior is observed, with random fluctuations, in all other cases, and also in the open-set scenario,
so we refrain from a tedious detailed analysis.
However, it is worth pointing out that, in the presence of compression,
the versions based on robust estimation (r-LRT and r-FLD) provide a consistent, and often significant, improvement over those relying on ML estimation.

\begin{table*}[]
	\setlength\tabcolsep{6pt}
	\centering
	\footnotesize
    \renewcommand{\ru}{\rule{0mm}{3mm}}
	\begin{tabular}{rr||c|c||c||c|c|c|c|c}
		\ru    d &   N & PRNU  &  NP   & ideal  &      SVM   &      LRT   &     r-LRT  &      FLD   &     r-FLD  \\ \hline\hline
		\ru      &   1 & 0.034 & 0.178 & 0.308  & \TB{0.189} &     0.130  &     0.165  &     0.173  & \TB{0.189} \\
		\ru   64 &  10 & 0.074 & 0.234 & 0.387  & \TA{0.273} &     0.202  &     0.247  &     0.217  & \TA{0.273} \\
		\ru      & 100 & 0.238 & 0.225 & 0.558  &     0.440  &     0.399  &     0.443  &     0.404  & \TA{0.458} \\ \hline
		\ru      &   1 & 0.108 & 0.219 & 0.424  & \TA{0.303} &     0.212  &     0.255  &     0.262  &     0.298  \\
		\ru  256 &  10 & 0.369 & 0.261 & 0.648  &     0.578  &     0.510  &     0.562  &     0.562  & \TA{0.579} \\
		\ru      & 100 & 0.738 & 0.268 & 0.877  &     0.845  &     0.823  &     0.833  & \TB{0.847} &     0.834  \\ \hline
		\ru      &   1 & 0.527 & 0.249 & 0.737  &     0.547  &     0.602  &     0.635  & \TB{0.664} &     0.647  \\
		\ru 1024 &  10 & 0.855 & 0.301 & 0.936  &     0.879  &     0.885  &     0.892  & \TB{0.900} &     0.873  \\
		\ru      & 100 & 0.948 & 0.313 & 0.974  & \TB{0.952} &     0.948  &     0.948  &     0.947  &     0.926  \\ \hline
	\end{tabular}
    \caption{Accuracy on the Dresden dataset in the closed-set scenario with compression (QF=90).
    All performance figures lower significantly with respect to uncompressed images.
    Fusion keeps providing large gains in the most challenging case.}
\label{tab:ACC_90}
\end{table*}

\begin{table*}[]
	\setlength\tabcolsep{6pt}
	\centering
	\footnotesize
    \renewcommand{\ru}{\rule{0mm}{3mm}}
	\begin{tabular}{rr||c|c||c||c|c|c|c|c}
		\ru    d &   N & PRNU  &  NP   & ideal  &      SVM   &      LRT   &     r-LRT  &      FLD   &     r-FLD  \\ \hline\hline
		\ru      &   1 & 0.558 & 0.859 & 0.979  & \TA{0.857} &     0.843  &     0.856  &     0.848  & \TA{0.857} \\
		\ru   64 &  10 & 0.642 & 0.897 & 0.982  &     0.897  &     0.887  & \TA{0.898} &     0.863  &     0.897  \\
		\ru      & 100 & 0.778 & 0.917 & 0.987  &     0.926  &     0.928  & \TB{0.934} &     0.910  &     0.933  \\ \hline
		\ru      &   1 & 0.689 & 0.891 & 0.984  & \TA{0.897} &     0.867  &     0.886  &     0.862  &     0.888  \\
		\ru  256 &  10 & 0.834 & 0.911 & 0.990  & \TB{0.933} &     0.925  &     0.929  &     0.929  & \TB{0.933} \\
		\ru      & 100 & 0.946 & 0.936 & 0.996  & \TB{0.971} &     0.967  &     0.967  & \TB{0.971} &     0.964  \\ \hline
		\ru      &   1 & 0.876 & 0.912 & 0.992  &     0.925  &     0.929  &     0.935  & \TB{0.939} &     0.937  \\
		\ru 1024 &  10 & 0.973 & 0.922 & 0.998  &     0.957  &     0.968  &     0.965  & \TB{0.970} &     0.954  \\
		\ru      & 100 & 0.990 & 0.946 & 0.999  &     0.983  &     0.981  &     0.980  & \TB{0.980} &     0.971  \\ \hline
	\end{tabular}
    \caption{AUC on the Dresden dataset in the open-set scenario with compression (QF=90).
    Note that in the most favourable conditions, fusion shows some small impairments with respect to PRNU-only classification.}
\label{tab:AUC_90}
\end{table*}

\begin{table*}[]
	\setlength\tabcolsep{6pt}
	\centering
	\footnotesize
    \renewcommand{\ru}{\rule{0mm}{3mm}}
	\begin{tabular}{rr||c|c||c||c|c|c|c|c}
		\ru    d &   N & PRNU  &  NP   & ideal  &      SVM   &      LRT   &     r-LRT  &      FLD   &     r-FLD  \\ \hline\hline
		\ru      &   1 & 0.027 & 0.141 & 0.299  &     0.137  &     0.086  &     0.130  &     0.120  & \TB{0.147} \\
		\ru   64 &  10 & 0.043 & 0.195 & 0.357  &     0.188  &     0.138  &     0.188  &     0.142  & \TB{0.217} \\
		\ru      & 100 & 0.149 & 0.219 & 0.482  &     0.295  &     0.271  &     0.324  &     0.248  & \TA{0.347} \\ \hline
		\ru      &   1 & 0.067 & 0.188 & 0.385  & \TB{0.228} &     0.146  &     0.193  &     0.162  &     0.218 \\
		\ru  256 &  10 & 0.229 & 0.246 & 0.568  &     0.448  &     0.352  &     0.426  &     0.356  & \TA{0.473} \\
		\ru      & 100 & 0.618 & 0.264 & 0.820  &     0.750  &     0.713  &     0.749  &     0.736  & \TB{0.772} \\ \hline
		\ru      &   1 & 0.364 & 0.232 & 0.643  &     0.475  &     0.436  &     0.485  &     0.482  & \TA{0.540} \\
		\ru 1024 &  10 & 0.774 & 0.283 & 0.899  & \TB{0.864} &     0.818  &     0.847  &     0.861  &     0.852  \\
		\ru      & 100 & 0.935 & 0.321 & 0.967  &     0.948  &     0.937  &     0.938  & \TB{0.948} &     0.930  \\ \hline
	\end{tabular}
    \caption{Accuracy on the Dresden dataset in the closed-set scenario with compression (QF=80).
    Now performance figures lower dramatically with respect to uncompressed images unless plenty of data ar available.
    Fusion keeps providing good gains, and almost always robust FLD perform much better than other classifiers.}
	\label{tab:ACC_80}
\end{table*}

\begin{table*}[]
	\setlength\tabcolsep{6pt}
	\centering
	\footnotesize
    \renewcommand{\ru}{\rule{0mm}{3mm}}
	\begin{tabular}{rr||c|c||c||c|c|c|c|c}
		\ru    d &   N & PRNU  &  NP   & ideal  &      SVM   &      LRT   &     r-LRT  &      FLD   &     r-FLD  \\ \hline\hline
		\ru      &   1 & 0.543 & 0.822 & 0.979  &     0.818  &     0.801  &     0.820  &     0.807  & \TA{0.820} \\
		\ru   64 &  10 & 0.609 & 0.867 & 0.981  &     0.852  &     0.848  &     0.868  &     0.807  & \TA{0.869} \\
		\ru      & 100 & 0.731 & 0.900 & 0.986  &     0.883  &     0.899  &     0.914  &     0.850  & \TA{0.911} \\ \hline
		\ru      &   1 & 0.639 & 0.862 & 0.982  & \TA{0.863} &     0.834  &     0.861  &     0.804  &     0.858  \\
		\ru  256 &  10 & 0.783 & 0.888 & 0.988  &     0.910  &     0.897  &     0.907  &     0.880  & \TB{0.913} \\
		\ru      & 100 & 0.920 & 0.921 & 0.995  &     0.959  &     0.956  &     0.959  &     0.956  & \TB{0.961} \\ \hline
		\ru      &   1 & 0.832 & 0.882 & 0.990  &     0.904  &     0.904  &     0.913  &     0.906  & \TB{0.918} \\
		\ru 1024 &  10 & 0.958 & 0.898 & 0.997  &     0.958  &     0.959  &     0.954  & \TB{0.971} &     0.950  \\
		\ru      & 100 & 0.986 & 0.932 & 0.999  & \TB{0.984} &     0.980  &     0.979  & \TB{0.984} &     0.972  \\ \hline
	\end{tabular}
    \caption{AUC on the Dresden dataset in the open-set scenario with compression (QF=80).
    Fusion preserves a reasonable performance even with scarce data, with robust FLD almost always the best method.}
	\label{tab:AUC_80}
\end{table*}

\section{Conclusions}
In this paper, we proposed to use noiseprint, a camera-model image fingerprint, to support PRNU-based forensic analyses.
Numerical experiments prove that the proposed approach ensures a significant performance improvement in several challenging situations easily encountered in real-world applications.

This is only a first step in this direction, and there is certainly much room for further improvements.
In future work we want to extend the proposed approach to improve PRNU-based image forgery detection and localization,
and also to perform accurate blind image clustering, an important problem in multimedia forensics.

\section{Acknowledgements}
This material is based on research sponsored by the Air Force Research Laboratory
and the Defense Advanced Research Projects Agency under agreement number
{\small FA8750-16-2-0204}. The U.S.Government is authorized to reproduce and distribute
reprints for Governmental purposes notwithstanding any copyright notation thereon.
The views and conclusions contained herein are those of the authors and should not
be interpreted as necessarily representing the official policies or endorsements,
either expressed or implied, of the Air Force Research Laboratory and the Defense
Advanced Research Projects Agency or the U.S. Government.

\balance

\bibliographystyle{spmpsci}      
\bibliography{refs}   

\end{document}